\PassOptionsToPackage{table,xcdraw,dvipsnames}{xcolor}
\documentclass{article}

\usepackage[dblblindworkshop, final]{neurips_2025}
\usepackage{graphicx}
\usepackage{tabularx} 
\usepackage{booktabs,multirow}

\usepackage[utf8]{inputenc} 
\usepackage[T1]{fontenc}    
\usepackage{hyperref}       
\usepackage{url}            
\usepackage{amsfonts}       
\usepackage{nicefrac}       
\usepackage{microtype}      
\usepackage{xcolor}         
\usepackage{graphicx}
\usepackage{wrapfig}
\usepackage{float}     
\usepackage{inconsolata} 
\usepackage{fdsymbol} 
\usepackage{bbm}

\definecolor{darkblue}{rgb}{0, 0, 0.5}
\hypersetup{colorlinks=true, citecolor=darkblue, linkcolor=darkblue, urlcolor=darkblue}

\title{Beyond Fertility: Analyzing STRR as a Metric for Multilingual Tokenization Evaluation}
\workshoptitle{Evaluating the Evolving LLM Lifecycle: Benchmarks, Emergent Abilities, and Scaling}

\author{%
Mir Tafseer Nayeem$^{\clubsuit}$\thanks{Equal contribution.} \enspace \enspace Sawsan Alqahtani$^{\spadesuit}$\footnotemark[1] \\ 
\bf Md Tahmid Rahman Laskar$^{\vardiamondsuit}$ \enspace Tasnim Mohiuddin$^{\diamondsuit}$ \enspace M Saiful Bari$^{\varheartsuit}$\\
$^\clubsuit$University of Alberta \enspace 
$^\spadesuit$Princess Nourah Bint Abdulrahman University\\
$^\vardiamondsuit$Dialpad \enspace
$^\diamondsuit$Qatar Computing Research Institute \enspace 
$^\varheartsuit$Amazon AGI
}

\begin{document}

\maketitle

\begin{abstract}
Tokenization is a crucial but under-evaluated step in large language models (LLMs). The standard metric, fertility (the average number of tokens per word) captures compression efficiency but obscures how vocabularies are allocated across languages and domains. We analyze six widely used tokenizers across seven languages and two domains, finding stable fertility for English, high fertility for Chinese, and little domain sensitivity. To address fertility’s blind spots, we propose the Single Token Retention Rate (STRR), which measures the proportion of words preserved as single tokens. STRR reveals systematic prioritization of English, strong support for Chinese, and fragmentation in Hindi, offering an interpretable view of cross-lingual fairness. Our results show that STRR complements fertility and provides practical guidance for designing more equitable multilingual tokenizers.\footnote{Our code and dataset are available at \href{https://github.com/tafseer-nayeem/STRR}{github.com/tafseer-nayeem/STRR}}

\end{abstract}

\section{Introduction}

Tokenization is a foundational step in large language models (LLMs), shaping how text is split into model-readable units, yet its evaluation remains under-examined and constrained by a lack of interpretable metrics \citep{bostrom2020byte}. Existing metrics often prioritize \textit{compression efficiency}, with \textit{fertility} (the average number of subword tokens generated per word) serving as a standard diagnostic \citep{rust2020good,ali2023tokenizer} (see \S\ref{sec:related_work} and \S\ref{app:tokenization} for other possible evaluation metrics). High fertility scores typically signal inefficiency, since more tokens are required to represent the same semantic content. Despite its wide adoption, fertility has important blind spots: as a token-level average, it obscures how vocabulary is allocated across languages, domains, and usage contexts, and its link to downstream LLM performance remains unclear \citep{bostrom2020byte}. Yet, as Table \ref{tab:languages-domains} suggests, fertility compresses behavior into a narrow numeric band and offers little diagnostic guidance about where vocabulary capacity is misallocated.

These limitations are consequential: tokenization governs how capacity is allocated, affecting downstream efficiency, fairness, and representation quality in LLMs. A tokenizer that fragments words in some languages more than others implicitly biases model capacity, inflating training and inference costs for those languages and amplifying performance disparities \citep{bostrom2020byte}. Moreover, evaluation centered solely on fertility obscures challenges that arise in multilingual and code-mixed scenarios, where speakers fluidly switch across linguistic boundaries \citep{JESR5628}. Such settings expose weaknesses in current tokenizers, particularly when English, functioning as a global lingua franca, interacts with diverse native languages \citep{english-lingua}. 

Despite advances in multilingual pretraining, tokenizers still struggle to balance two competing goals: preserving coverage across diverse languages, and scripts, while minimizing fragmentation. We argue that existing evaluation practices are insufficient to guide tokenizer design toward this balance. 

To address this gap, we contribute in two ways. First, we present a cross-lingual evaluation of six LLM tokenizers across seven languages and two domains (formal and informal)~(\S\ref{sec:fertility_analysis}). Second, we introduce the Single Token Retention Rate (STRR), a novel metric that measures the proportion of words preserved as single tokens across languages. Unlike fertility, STRR better captures tokenizers’ vocabulary allocation and provides an interpretable diagnostic for fairness and efficiency (\S\ref{ssec:STRR}). Together, these analyses shed light on how contemporary tokenizers implicitly prioritize certain languages and suggest directions for equitable and efficient multilingual tokenizer design~(\S\ref{sec:discussions}).

\section{Related Work}
\label{sec:related_work}
Most tokenizer evaluations rely on \textit{fertility} (the average number of tokens per word) valued for its simplicity but limited in scope \citep{rust2020good}. Other measures such as vocabulary coverage, subword entropy, compression rates, or character-to-token ratios have been proposed \citep{goldman2024unpacking,zouhar2023tokenization,libovicky2024lexically,signoroni-rychly-2022-hft,lotz2025beyond}, yet none have become standard practice. Linguistically motivated metrics also exist \citep{arnett2025evaluating,beinborn2023analyzing,asgari2025morphbpemorphoawaretokenizerbridging}, but they are often language-specific and difficult to interpret across diverse scripts.

These approaches emphasize compression efficiency but rarely reveal how vocabulary is distributed across languages or domains, nor do they consistently correlate with downstream model performance~\citep{bostrom2020byte,ali2023tokenizer}. Prior work highlights the consequences of uneven token allocation: inflated inference costs for some languages~\citep{ahia-etal-2023-languages}, reduced cross-domain robustness~\citep{dagan2024getting}, and misalignment with linguistic boundaries~\citep{yin2024compositional,bogin2022unobserved}. Together, these studies underscore the need for standardized, interpretable metrics that capture both efficiency and fairness in multilingual settings.

\section{Experimental Setup}
\label{sec:exp_setup}
\textbf{Tokenizers:} We selected six widely used LLM tokenizers: \texttt{GPT-4o},
\texttt{Aya-Expanse-32B} ~\citep{dang2024ayaexpansecombiningresearch}, \texttt{Mistral-Small-24B}\footnote{\url{https://mistral.ai/en/news/mistral-small-3}}, \texttt{Llama-3.1-70B}~\citep{dubey2024llama3herdmodels}, \texttt{Qwen2.5-72B}~\citep{qwen2025qwen25technicalreport}, and \texttt{DeepSeek-V3}~\citep{deepseekai2024deepseekv3technicalreport}.


\paragraph{Datasets:} For fertility and related metrics we use formal text (XL-Sum news; \citealt{hasan-etal-2021-xl}) and informal text (MultilingualSentiment; \citealt{clapAI2024multilingualsentiment}). 
For STRR, we build a multilingual wordlist from \texttt{1000MostCommonWords}\footnote{\url{https://1000mostcommonwords.com}}, aligning 1,000 translation pairs (e.g., English–French) per language to ensure cross-lingual comparability and reflect high-frequency vocabulary.

\textbf{Languages:} We consider several languages (English, German, French, Spanish, Italian, Hindi, and Chinese) selected because they are (i) officially supported by the evaluated LLMs and (ii) included in widely used multilingual benchmarks (e.g., MMMLU\footnote{\url{https://huggingface.co/datasets/openai/MMMLU}}). For the fertility analyses, we restrict to English, French, Spanish, and Chinese to ensure uniform data availability across both formal and informal domains in the chosen datasets.

\section{Fertility Analysis}
\label{sec:fertility_analysis}

Table~\ref{tab:languages-domains} reports fertility scores across languages and tokenizers. Additional metrics, subword entropy and characters-per-token (defined in \S\ref{app:tokenization}), are shown in Table~\ref{tab:all_in_one}, and display trends consistent with the fertility results.
English shows striking consistency across both formal and informal domains, reflecting its dominance in pretraining corpora~\citep{dubey2024llama3herdmodels,deepseekai2024deepseekv3technicalreport} and relatively simple morphology~\citep{bentz-etal-2016-comparison}. 


\begin{wraptable}{r}{0.49\textwidth}
\centering
\normalsize
\setlength{\tabcolsep}{4pt}
\renewcommand{\arraystretch}{1.1}
\resizebox{0.49\textwidth}{!}{
\begin{tabular}{cccccccc}
\hline
\textbf{Languages} & \textbf{Domains} & \textbf{GPT} & \textbf{Aya-exp} & \textbf{Mistral} & \textbf{Llama} & \textbf{Qwen} & \textbf{DeepSeek} \\ 
& & \textbf{4o} & \textbf{32B} & \textbf{24B} & \textbf{3.1-70B} & \textbf{2.5-72B} & \textbf{V3} \\
\hline \hline
\multirow{2}{*}{\textbf{English}} & Formal   & 1.22 & 1.24 & 1.27 & 1.23 & 1.25 & 1.23 \\
                                  & Informal & 1.22 & 1.25 & 1.27 & 1.25 & 1.26 & 1.25 \\ \hline
\multirow{2}{*}{\textbf{French}}  & Formal   & 1.42 & 1.42 & 1.43 & 1.67 & 1.68 & 1.61 \\
                                  & Informal & 1.37 & 1.42 & 1.41 & 1.58 & 1.58 & 1.57 \\ \hline
\multirow{2}{*}{\textbf{Spanish}} & Formal   & 1.36 & 1.33 & 1.42 & 1.61 & 1.61 & 1.55 \\
                                  & Informal & 1.32 & 1.36 & 1.44 & 1.53 & 1.53 & 1.53 \\ \hline
\multirow{2}{*}{\textbf{Chinese}} & Formal   & 1.89 & 1.82 & 2.21 & 1.89 & 2.40 & 1.95 \\
                                  & Informal & 1.86 & 1.96 & 2.30 & 1.92 & 2.40 & 1.95 \\ \hline
\end{tabular}
}
\vspace{-7pt}
\caption{Fertility values across languages, domains (formal and informal), and tokenizers.}
\label{tab:languages-domains}
\vspace{-9pt}
\end{wraptable}

Overall, domain differences are minimal, suggesting that large vocabularies (128K–255K; Appendix, Table~\ref{tab:model-details}) capture both structured and unstructured text efficiently. In contrast, Chinese exhibits the highest fertility due to its logographic script and absence of explicit word boundaries~\citep{si-etal-2023-sub}. Tokenizers vary in how much vocabulary they allocate to whole words versus smaller units; fertility reflects these numerical differences but cannot distinguish necessary linguistic segmentation from suboptimal allocation.

While informative, fertility, subword entropy, and characters-per-token each have blind spots that limit their usefulness for equitable multilingual tokenizer design. Fertility collapses behavior into average tokens per word, masking over-fragmentation. Subword entropy summarizes distributional balance but remains abstract and hard to localize. Characters-per-token highlights script differences but reduces quality to mean token length, ignoring whether frequent words remain intact.

\section{Single Token Retention Rate (STRR)}
\label{ssec:STRR}
We propose the Single-Token Retention Rate (STRR), which measures the proportion of words preserved as single tokens. STRR serves two goals: (i) probing vocabulary construction by quantifying whole-word retention in each language, and (ii) revealing how tokenizers allocate limited vocabulary across languages. It highlights inequities directly and points to actionable remedies, such as expanding coverage for under-represented high-frequency words. Unlike fertility, subword entropy, or characters-per-token—which are computed on text corpora and averaged over tokenized outputs; STRR is defined on a reference wordlist. It checks, for each word, whether the tokenizer has allocated a single token, making it a type-level rather than token-level diagnostic. This design makes STRR interpretable, fairness-sensitive, and tied to practical interventions.



\subsection{Definition}
Given a set of words \(W=\{w_1,\dots,w_n\}\) and a tokenizer \(T\), we define
\vspace{-0.1em}
\[
\mathrm{STRR}(T;W)=\frac{1}{n}\sum_{i=1}^n \mathbbm{1}\!\left(|T(w_i)|=1\right)\times 100;
\]
\vspace{-0.1em}
STRR thus measures the percentage of words encoded as a single token.

\subsection{Results}
\begin{figure*}[t]
    \centering
    \includegraphics[scale = 0.37]{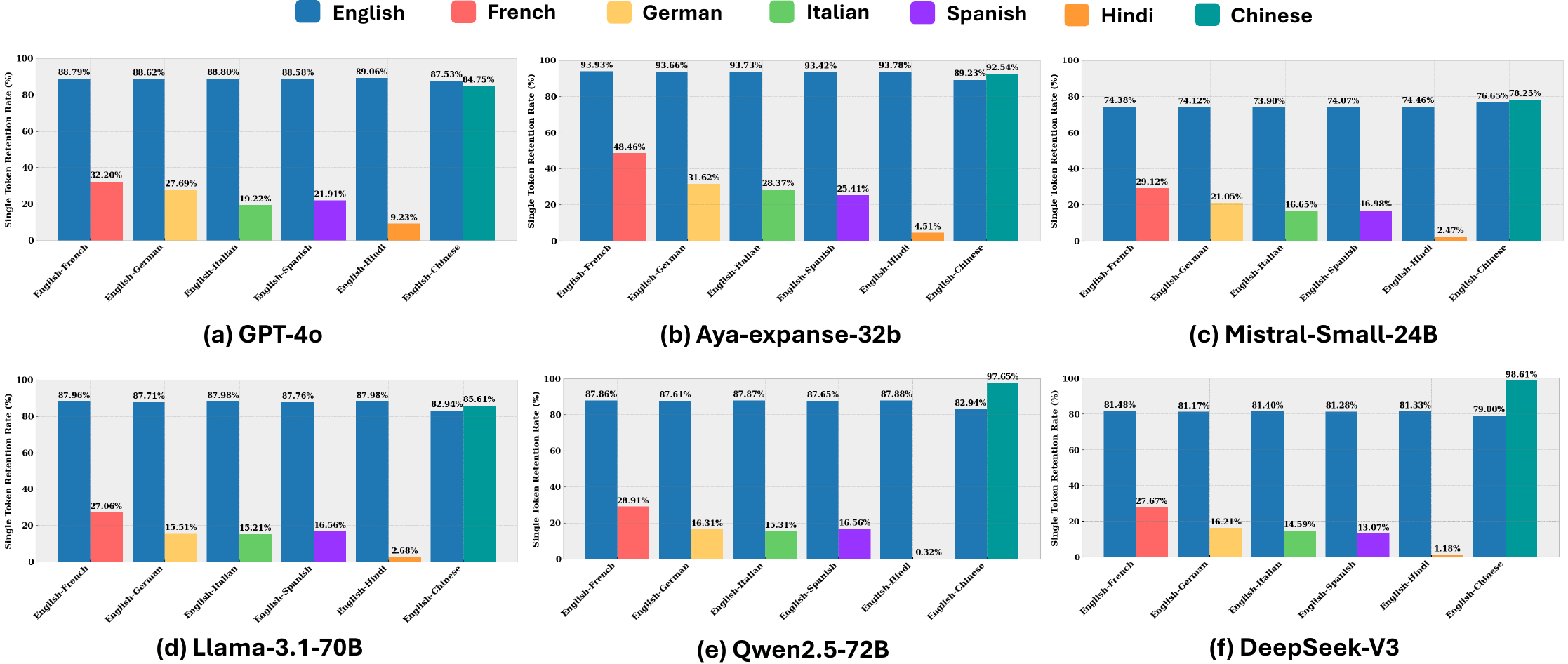}
    \caption{Single Token Retention Rate (STRR) across six LLM tokenizers for different language pairs. Each pair (e.g., English-French) represents 1,000 parallel words in both languages, allowing us to examine whether LLM tokenizers prioritize the English versions of words over their multilingual counterparts. A high STRR for English suggests that tokenizers allocate more vocabulary space to English, 
    while differences in STRR across languages indicate varying degrees of support.
    }
    \label{fig:STRR}
\end{figure*}

As illustrated in Figure~\ref{fig:STRR}, across all tokenizers, \textit{\textbf{English words in translation pairs are overwhelmingly retained as single tokens}}. This supports the hypothesis that tokenizers allocate significant vocabulary space to English representations, reinforcing findings that even limited multilingual exposure enhances LLM multilingual capabilities~\citep{shaham-etal-2024-multilingual}, as models primarily learn direct mappings from English tokens to multilingual equivalents, reducing reliance on extensive multilingual pretraining.

Our STRR analysis reveals that \textit{\textbf{all LLMs explicitly integrate Chinese vocabulary into their tokenization strategies}} to reduce segmentation artifacts as observed in Table~\ref{tab:languages-domains}. Notably, Qwen2.5-72B and DeepSeek-V3 exhibit the highest STRR for Chinese, suggesting enhanced language-specific support for whole-word representations.

\textit{\textbf{Hindi exhibits the lowest STRR across all evaluated tokenizers}}, revealing pronounced fragmentation and suboptimal vocabulary allocation. Crucially, STRR quantifies this inefficiency with a direct, interpretable measure, rather than simply echoing prior fertility-based findings~\citep{ahia-etal-2023-languages}, offering clear guidance for targeted vocabulary expansion in under‑served languages (\S\ref{sec:discussions}).

\section{Discussion \& Recommendations}
\label{sec:discussions}

\paragraph{Identifying Core Vocabulary via the Pareto Principle:} The Pareto Principle, or “80/20 rule,” posits that a small fraction of the lexicon accounts for most language use~\citep{Sanders1987Pareto}.  In English, the General Service List (GSL) of roughly 2,000 words covers 80–85\% of standard written text~\citep{West1953GSL}.  We thus advocate that multilingual tokenizer developers identify an analogous \emph{core vocabulary} in each language (namely, the highest‐frequency words that dominate token counts) and ensure they are encoded as single tokens.  Prioritizing this compact set minimizes subword fragmentation and maximizes encoding efficiency without unnecessarily expanding the overall vocabulary.

\paragraph{End‐to‐End Vocabulary Expansion Pipeline:}  
We propose a practical four‐stage pipeline for enhancing multilingual tokenizers, feasible even in low‐resource settings or without large pretraining corpora. As a shared baseline, we release curated lists of the 1,000 most frequent words in seven major languages. 

\begin{enumerate}
  \item \textbf{Core Vocabulary Identification:}  
    Select the highest‐frequency words in each target language using our curated lists or extend them as needed.\footnote{\url{https://1000mostcommonwords.com/languages/}}  

  \item \textbf{Vocabulary Injection:}  
    Add identified words to the tokenizer’s vocabulary as single tokens. Use STRR to check which are already represented and which require injection (\S\ref{ssec:STRR}).  

  \item \textbf{Corpus Pretraining:}  
    Continue pretraining or fine‐tuning the base multilingual LLM on publicly available multilingual text~\citep{ustun-etal-2024-aya}, incorporating the expanded vocabulary to learn robust embeddings.  

  \item \textbf{Multilingual Instruction Tuning:}  
    Instruction‐tune the model on multilingual instruction–response datasets~\citep{singh-etal-2024-aya} to validate and reinforce the expanded vocabulary in downstream tasks.  
\end{enumerate}

This pipeline can reduce subword fragmentation, facilitate faster adaptation, and potentially improve consistency across diverse languages.

\section{Conclusion}
We introduced STRR, a simple interpretable metric that complements fertility by capturing whole-word preservation in multilingual tokenization. Our analysis across tokenizers shows that STRR reveals biases, favoring English and Chinese while fragmenting languages like Hindi, that fertility alone cannot. By releasing high-frequency word lists, providing code, and outlining a vocabulary-expansion pipeline, we offer actionable steps toward more efficient and equitable tokenizer design. 

\bibliography{custom}
\bibliographystyle{unsrtnat}
{
}
\appendix

\section{Model Details}
\begin{table}[ht]
\centering
\renewcommand{\arraystretch}{1.1} 
\setlength{\tabcolsep}{4pt} 
\resizebox{10cm}{!} 
{
\begin{tabular}{ccc}
\hline
\rowcolor[HTML]{EFEFEF} 
\textbf{Models}   & \textbf{Vocab Size} & \textbf{Model ID} \\ \hline \hline
\href{https://openai.com/index/hello-gpt-4o/}{\texttt{GPT-4o}}             & 200,019             & \href{https://openai.com/index/hello-gpt-4o/}{Link}             \\
\texttt{Aya-Expanse-32B} (\cite{dang2024ayaexpansecombiningresearch})   & 255,029            & \href{https://huggingface.co/CohereForAI/aya-expanse-32b}{HF Link}              \\
\href{https://mistral.ai/en/news/mistral-small-3}{\texttt{Mistral-Small-24B}} & 131,072             & \href{https://huggingface.co/mistralai/Mistral-Small-24B-Instruct-2501}{HF Link}              \\
\texttt{Llama-3.1-70B} (\cite{dubey2024llama3herdmodels})     & 128,256             & \href{https://huggingface.co/meta-llama/Llama-3.1-70B-Instruct}{HF Link}             \\
\texttt{Qwen2.5-72B} (\cite{qwen2025qwen25technicalreport})       & 151,665             & \href{https://huggingface.co/Qwen/Qwen2.5-72B-Instruct}{HF Link}             \\
\texttt{DeepSeek-V3} (\cite{deepseekai2024deepseekv3technicalreport})      & 128,815             & \href{https://huggingface.co/deepseek-ai/DeepSeek-V3}{HF Link}            \\ \hline
\end{tabular}
}
\caption{Details of the models used in our experiments, including total vocabulary size (with added tokens) for each model. "HF Link" refers to the corresponding Hugging Face model IDs.}
\label{tab:model-details}
\end{table}

\section{Tokenization Metrics: Definitions and Limitations}
\label{app:tokenization}
\begin{table}[!htbp]
\centering
\renewcommand{\arraystretch}{1.3}
\begin{tabularx}{\columnwidth}{|p{2.5cm}|X|X|X|}
\hline
\textbf{Metric} & \textbf{Definition} & \textbf{What it Captures} & \textbf{How STRR Differs} \\
\hline
\textbf{Fertility} 
& Avg.\ number of tokens per word (compression proxy). 
& Measures sequence length efficiency; high fertility = more fragmentation. 
& STRR is type-level: counts \% of whole words preserved. Fertility hides where fragmentation occurs, STRR pinpoints cross-lingual allocation. \\
\hline
\textbf{Subword Entropy} 
& Entropy of token frequency distribution across text. 
& Captures balance of vocabulary usage (skew vs.\ uniformity). High entropy = fairer, balanced allocation. 
& STRR measures whole-word retention per language, not distribution balance. Entropy flags global skew; STRR identifies which languages’ words are fragmented. \\
\hline
\textbf{Char-to-Token Ratio} 
& Avg.\ number of characters per token. 
& Captures average token length; highlights script differences (e.g., Chinese vs.\ English). 
& STRR does not average token lengths, but directly counts intact words. Differentiates many slightly split words from severe fragmentation of core vocabulary. \\
\hline
\textbf{STRR (Single Token Retention Rate)} 
& Percentage of words preserved as single tokens. 
& Captures vocabulary allocation fairness and whole-word coverage across languages. 
& Provides actionable, interpretable diagnostic: directly shows which languages and words are under-served and can guide vocabulary expansion. \\
\hline
\end{tabularx}
\caption{Comparison of tokenization evaluation metrics. Fertility and char-to-token ratio measure compression/fragmentation averages; subword entropy measures distributional balance; STRR highlights cross-lingual fairness by directly quantifying whole-word retention.}
\label{tab:metrics_comparison}
\end{table}


\begin{table}[!htbp]
\centering
\small
\label{tab:all_in_one}
\resizebox{\textwidth}{!}{
\begin{tabular}{ll ccc ccc ccc ccc ccc ccc}
\toprule
\multirow{2}{*}{\textbf{Language}} & \multirow{2}{*}{\textbf{Domain}} &
\multicolumn{3}{c}{\textbf{GPT-4o}} &
\multicolumn{3}{c}{\textbf{Aya-Expanse-32B}} &
\multicolumn{3}{c}{\textbf{Mistral-Small-24B}} &
\multicolumn{3}{c}{\textbf{Llama-3.1-70B}} &
\multicolumn{3}{c}{\textbf{Qwen2.5-72B}} &
\multicolumn{3}{c}{\textbf{DeepSeek-V3}} \\
\cmidrule(lr){3-5}\cmidrule(lr){6-8}\cmidrule(lr){9-11}\cmidrule(lr){12-14}\cmidrule(lr){15-17}\cmidrule(lr){18-20}
& & \textbf{Fert.} & \textbf{Ent.} & \textbf{Chars/Tok} & \textbf{Fert.} & \textbf{Ent.} & \textbf{Chars/Tok} & \textbf{Fert.} & \textbf{Ent.} & \textbf{Chars/Tok} & \textbf{Fert.} & \textbf{Ent.} & \textbf{Chars/Tok} & \textbf{Fert.} & \textbf{Ent.} & \textbf{Chars/Tok} & \textbf{Fert.} & \textbf{Ent.} & \textbf{Chars/Tok} \\
\midrule
\multirow{2}{*}{English} & Formal   & 1.22 & 9.45 & 3.88 & 1.24 & 9.30 & 3.80 & 1.27 & 9.40 & 3.72 & 1.23 & 9.43 & 3.84 & 1.25 & 9.36 & 3.77 & 1.23 & 9.44 & 3.83 \\
                         & Informal & 1.22 & 9.59 & 3.76 & 1.25 & 9.49 & 3.67 & 1.27 & 9.61 & 3.59 & 1.25 & 9.59 & 3.66 & 1.26 & 9.57 & 3.64 & 1.25 & 9.61 & 3.65 \\
\midrule
\multirow{2}{*}{French}  & Formal   & 1.42 & 9.75 & 3.60 & 1.42 & 9.44 & 3.60 & 1.43 & 9.70 & 3.58 & 1.67 & 9.70 & 3.06 & 1.68 & 9.67 & 3.06 & 1.61 & 9.73 & 3.18 \\
                         & Informal & 1.37 & 9.92 & 3.54 & 1.42 & 9.74 & 3.43 & 1.41 & 9.93 & 3.45 & 1.58 & 9.85 & 3.06 & 1.58 & 9.85 & 3.08 & 1.57 & 9.85 & 3.09 \\
\midrule
\multirow{2}{*}{Spanish} & Formal   & 1.36 & 9.64 & 3.73 & 1.33 & 9.55 & 3.82 & 1.42 & 9.64 & 3.60 & 1.61 & 9.67 & 3.16 & 1.61 & 9.65 & 3.16 & 1.55 & 9.66 & 3.29 \\
                         & Informal & 1.32 & 9.32 & 3.50 & 1.36 & 9.30 & 3.39 & 1.44 & 9.33 & 3.22 & 1.53 & 9.32 & 3.01 & 1.53 & 9.32 & 3.02 & 1.53 & 9.31 & 3.01 \\
\midrule
\multirow{2}{*}{Chinese} & Formal   & 1.89 & 9.02 & 0.90 & 1.82 & 7.56 & 0.93 & 2.21 & 8.14 & 0.77 & 1.89 & 9.29 & 0.89 & 2.40 & 8.25 & 0.71 & 1.95 & 6.64 & 0.87 \\
                         & Informal & 1.86 & 8.55 & 0.84 & 1.96 & 7.03 & 0.80 & 2.30 & 7.68 & 0.68 & 1.92 & 8.78 & 0.82 & 2.40 & 7.89 & 0.65 & 1.95 & 6.30 & 0.80 \\
\bottomrule
\end{tabular}
}
\caption{Complete results: fertility (tokens/word), entropy (bits), and characters per token across languages, domains, and models.}
\end{table}

\end{document}